\documentclass{bmvc2k}

\def\ie{\textit{i.e.}}
\def\eg{\textit{e.g.}}

\def\etal{\textit{et al.~}}

\usepackage{graphicx}
\usepackage{float}
\usepackage{subfigure}
\usepackage{caption}
\usepackage{colortbl} % \cellcolor
\usepackage{amsmath}
\usepackage{amssymb}
\usepackage{tabularx}
\usepackage{multirow}
\usepackage{color,xcolor}

\newcommand{\Mat}[1]{\mathbf{#1}}

\newcommand{\wrt}{\emph{w.r.t.}}
\newcommand{\cf}{\emph{cf.}}

\definecolor{mygray-bg}{gray}{0.9}

\usepackage[font=small]{caption}

\title{Rethinking the Evaluation of Unbiased \\ Scene Graph Generation}

\addauthor{Xingchen Li}{xingchenl@zju.edu.cn}{1}
\addauthor{Long Chen$^\dagger$}{zjuchenlong@gmail.com}{2}
\addauthor{Jian Shao}{jshao@zju.edu.cn}{1}
\addauthor{Shaoning Xiao}{shaoningx@zju.edu.cn}{1}
\addauthor{Songyang Zhang}{szhang83@ur.rochester.edu}{3}
\addauthor{Jun Xiao}{junx@zju.edu.cn}{1}

\addinstitution{
 College of Computer Science \\
 Zhejiang University \\
 Hangzhou, China
}
\addinstitution{
 Department of Electrical Engineering \\
 Columbia University \\
 New York, USA \\
 $^\dagger$ Corresponding author
}
\addinstitution{
 Department of Computer Science \\
 University of Rochester \\
 Rochester, USA \\
}

\runninghead{Li, Chen, Shao \etal}{Rethinking the Evaluation of Unbiased SGG}

\def\eg{\emph{e.g}\bmvaOneDot}

\def\etal{\emph{et al}\bmvaOneDot}

\begin{document}

\maketitle

\begin{abstract}
    Current Scene Graph Generation (SGG) methods tend to predict frequent predicate categories and fail to recognize rare ones due to the severe imbalanced distribution of predicates.
    To improve the robustness of SGG models on different predicate categories, recent research has focused on unbiased SGG and adopted \emph{mean Recall@K} (mR@K) as the main evaluation metric.
    However, we discovered two overlooked issues about this de facto standard metric, which makes current unbiased SGG evaluation vulnerable and unfair:
    1) mR@K neglects the correlations among predicates and unintentionally breaks category independence when ranking all the triplet predictions together regardless of the predicate categories.
    2) mR@K neglects the compositional diversity of different predicates and assigns excessively high weights to some oversimple category samples with limited composable relation triplet types. 
    In addition, we investigate the under-explored correlation between objects and predicates, which can serve as a simple but strong baseline for unbiased SGG. 
    In this paper, we refine mR@K and propose two complementary evaluation metrics for unbiased SGG: Independent Mean Recall (\textbf{IMR}) and weighted IMR (\textbf{wIMR}). These two metrics are designed by considering the category independence and diversity of composable relation triplets, respectively. We compare the proposed metrics with the de facto standard metrics through extensive experiments and discuss the solutions to evaluate unbiased SGG in a more trustworthy way\footnote{Code will be available at \url{https://github.com/xcppy/Unbiased_SGG}.}.
\end{abstract}

%-------------------------------------------------------------------------
\section{Introduction}

Scene graphs are prevailing visually-grounded graph structured representations for scene understanding, which consist of a set of visual relation triplets (\ie, $\langle$\texttt{subject}, \texttt{predicate}, \texttt{object}$\rangle$)~\cite{cvpr17:IMP+,nips17:Pixel2Graph,yang2018graph}. 
Given an input image, Scene Graph Generation (SGG) requires models to not only localize and classify object categories accurately but also infer visual relationships (\ie, predicates) between pairwise objects.
Due to the inherent interpretability of scene graphs, they have been widely used in many downstream vision-and-language tasks, such as image retrieval~\cite{johnson2015image, wang2020cross, schroeder2020structured}, image captioning~\cite{gu2019unpaired, li2019know,yang2019auto} and visual grounding~\cite{jing2020visual}.

\begin{figure}[t]
    \centering
    \setlength{\abovecaptionskip}{0.28cm}
    \includegraphics[width=1.0\linewidth]{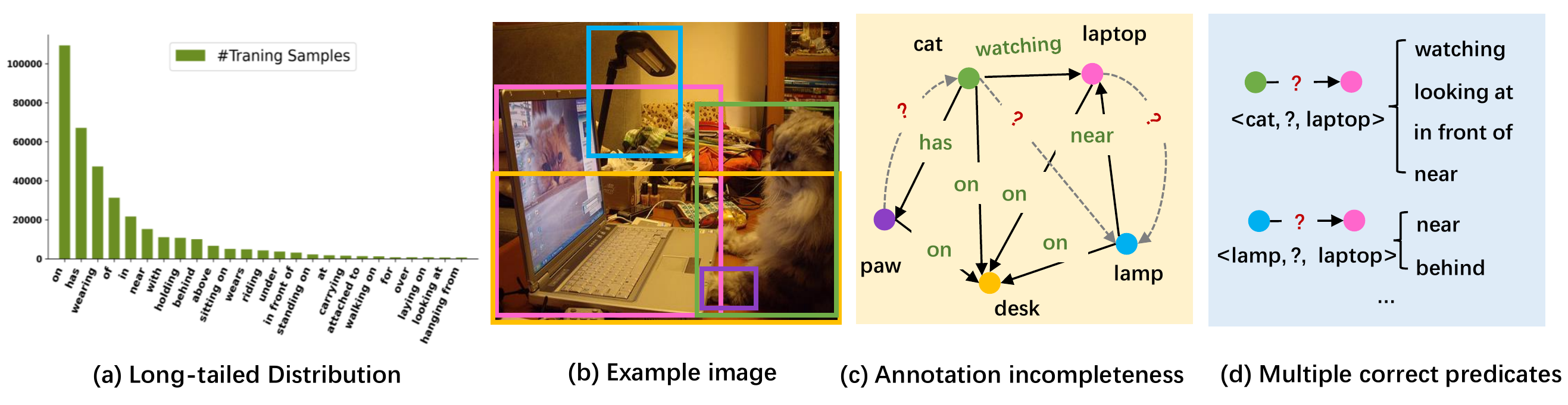}
    \vspace{-2em}
    \caption{Illustration of annotation characteristics of the VG dataset.  (\textbf{a}) The number distribution of samples for 25 most frequent predicates in VG. (\textbf{c}) The annotated ground-truths of the example image (b) and some missing relation triplets. (\textbf{d}) Some reasonable predicates for unannotated object pairs.}
    \label{fig:intro}
    
\end{figure}

However, it is hard to properly evaluate SGG models due to the intrinsic complexity of the task and several inevitable annotation characteristics of SGG datasets (\eg visual genome (VG)~\cite{visualGenome}):
1) The annotations of SGG datasets are \textbf{\emph{highly incomplete}}, \ie, it is unrealistic to exhaustively annotate all reasonable visual relationship triplets in one image, thus lots of positive triplets are missing. For example in Fig.~\ref{fig:intro}~(c), $\langle$\texttt{cat}, \texttt{near}, \texttt{lamp}$\rangle$ and $\langle$\texttt{paw}, \texttt{of}, \texttt{cat}$\rangle$ are missing in ground-truth annotations.
To avoid falsely penalizing the unannotated positive visual relations, early SGG work~\cite{qi2019attentive, wang2019exploring, li2018factorizable} exploits Recall@K (R@K) as the evaluation metric rather than other prevalent metrics (\eg, mean average precision, mAP), 
which measures the fraction of ground-truth visual relationship triplets that appear among the top-K most confident triplet predictions in the image.
2) The data distribution of predicate categories is \textbf{\emph{long-tailed}} (\cf~Fig.~\ref{fig:intro}(a)), \ie, ``head'' predicates\footnote{For conciseness, we use ``head'' (or ``tail'') predicates to represent these predicate categories in the ``head'' (or ``tail'') part of this long-tailed distributions in the following sections.} always have much more samples than ``tail'' ones, leading to severe bias for model evaluation --- the predicates for the head categories dominate R@K performance.
To observe the performance on each predicate category and treat each category independently, recent work~\cite{cvpr19:KERN, kaihuaCausal,cvpr21:BGNN,wang2020memory,PCPL} follow a new metric for evaluating unbiased SGG --- mean Recall@K (mR@K).

Specifically, the implementation of mR@K can be summarized into three steps: 
1) For each image, it first ranks top-K triplet predictions as outputs according to the confidence scores of all triplets and calculates the recall scores on each predicate category separately.
2) Then, it calculates the recall on each predicate category by averaging the corresponding recall scores over all images.
3) Finally, the recall scores for all predicate categories are averaged together to obtain the mR@K. 
\textbf{\emph{However, there are still two overlooked critical issues that make mR@K vulnerable and unfair:}}

The first issue is that the default mR@K unintentionally breaks the \emph{category independence} when ranking across categories to output top-K predictions, \ie, all triplet predictions are ranked together by their confidence scores regardless of their predicate categories. 
By ``category independence'', we mean that we hope the evaluation of each category should not be influenced by other categories. 
The setting of ranking across categories has no problem applying to other conventional single-label classification paradigms. Unfortunately, in SGG dataset, the predicates are highly correlated and there might be multiple reasonable predicates to describe the relationship for one object pair~\cite{li2022devil,li2022nicest}.
For example in Fig.~\ref{fig:intro}~(d), predicate \texttt{near} and \texttt{behind} are both correct for the subject-object pair $\langle$\texttt{lamp}, \texttt{laptop}$\rangle$. However, SGG is annotated following a single-label paradigm and only \texttt{near} is regarded as the ground-truth\footnote{Following the conventions in previous SGG work, for each subject-object pair, only one ground-truth predicate is used for the training and evaluation.}.
In such a case, the predictions of those predicates that have high correlations with other predicates will receive lower confidence scores after the normalization for single-label classification (\eg, Softmax). They are then less likely to appear in the top-K when ranking with the predictions of other predicates with low correlations,
leading to underestimated recall scores for those categories.
Since each predicate category in the dataset has different correlations with other categories, it is unfair to rank their predictions together.

The second issue is that the standard mR@K assigns equal weights to all predicate categories and neglects the \emph{compositional diversity} of predicates in subject-object categories.
By ``compositional diversity'', we mean the range of applicable scenarios for a predicate type, which can be measured by the number of possible compositions of subject-object category pairs.
In existing SGG datasets, the compositional diversity of different predicates varies greatly.
Specifically, in VG dataset, the predicate \texttt{on} has more than $4K$ types of subject-object category pairs while predicate \texttt{flying-in} only has two types.
Meanwhile, we observe that the tail predicates with fewer samples in SGG dataset usually possess limited composable relation triplets. Thus, mR@K, by assigning equal weights to each category, actually assigns excessively high weights to the samples of these tail predicates.
However, SGG is a task to understand rich visual semantics of images, which encourages models to recognize more types of visual relationship triplets.
The simple category-wise averaging strategy can not reflect this goal of the task.
And our experiments found that a simple baseline could easily trick mR@K by blindly utilizing the subject-object priors of those predicates with limited compositional diversity. 

In addition, to further benchmark unbiased SGG methods, we investigate the intrinsic correlation between objects and predicates from a new perspective. Different from previous frequency bias~\cite{motifs}, we turn our attention to the distribution of object categories under each predicate category and find this statistical prior can better reflect the correlation between objects and predicates. Based on this, we devise a simple baseline by directly aggregating this statistical prior into the predictions of SGG models and it greatly improves unbiased SGG performance. We think that this simple but strong method can serve as a baseline for the following unbiased SGG research.

In this paper, we refined the default mR@K and proposed two complementary metrics --- Independent Mean Recall (\textbf{IMR}) and weighted IMR (\textbf{wIMR}) --- to help evaluate unbiased SGG more fairly. 
Specifically, IMR ranks the predictions for each category independently to remove the mutual influence between categories,
and wIMR assigns different weights to each predicate category by considering compositional diversity.
We evaluate plenty of state-of-the-art unbiased SGG models on both current and new proposed metrics to compare their difference and discuss the solutions to evaluate unbiased SGG in a more trustworthy way.

\section{Related Work}

\noindent\textbf{Scene Graph Generation.}
The mainstream SGG methods can be classified into two major categories: 1) Two-stage pipeline~\cite{vcTree, motifs,cvpr19:KERN,PCPL,cogtree,li2022label}, which detect the objects first and then output the relationship predictions for every object pairs; 2) One-stage pipeline~\cite{liu2021fully, nips17:Pixel2Graph, liao2020ppdm}, which simultaneously detect objects and relationships.
Most SGG studies are based on the two-stage pipeline~\cite{chen2019scene,liang2019vrr,li2017scene,zhou2019visual,tian2020part}.  
Some work focused on learning better contextual features~\cite{cvpr17:IMP+,motifs,vcTree}.
They encode visual context of images into predicate representations by exploiting various powerful context encoding architectures, such as Bi-LSTM~\cite{motifs}, TreeLSTM~\cite{vcTree}, and graphs~\cite{tian2020part,cvpr17:IMP+, cvpr21:BGNN, mi2020hierarchical}. 
In addition, some studies are interested in improving the training loss~\cite{chen2019counterfactual}.
Aside from these two-stage methods,
there are also some other studies focusing on one-stage pipeline~\cite{liu2021fully, nips17:Pixel2Graph, liao2020ppdm, kim2020uniondet}.
Compared to two-stage methods, they are trained in an end-to-end manner. Typically, they usually treat the detection of objects and relationships as a point detection problem. 
For instance, in PPDM~\cite{liao2020ppdm} and IPNet~\cite{wang2020learning}, they separately localize and classify the object points and interaction points, and then group them.
Besides, since the fully supervised SGG requires excessive human labeling efforts, some recent work~\cite{zhong2021learning, li2022integrating} begins to focus on weakly supervised SGG. They only utilized image-caption pairs to train SGG models. The main challenge of weakly supervised SGG is to generate an alignment between text entities in captions and image regions.

\noindent\textbf{Evaluation Metrics for SGG.}
SGG evaluation is always challenging due to the intrinsic complexity of the SGG tasks and annotation characteristics. 
It is continuously discussed in many SGG studies~\cite{eccv16:VL,nips17:Pixel2Graph,vcTree,cvpr19:KERN,knyazev2020graph}. 
In early work~\cite{eccv16:VL,cvpr17:IMP+,nips17:Pixel2Graph,motifs}, the most widely used metric is Recall@K (R@K).
It measures the fraction of the ground-truths that appear among the top-$K$ most confident predictions in an image. 
In ~\cite{nips17:Pixel2Graph}, they allowed each subject-object pair to have multiple predicates, which means all the predicates will be involved in the recall ranking for each subject-object pair not just the one with the highest score. 
This setting significantly improves the performance of R@K and is named No Graph Constraint Recall@K (ngR@K).
In~\cite{knyazev2020graph}, they propose Weighted Triplet Recall@K to calculate recall by averaging the recall score on each type of visual relationship triplet.   
Besides, increasing recent studies have realized the long-tailed issue in SGG datasets, they propose Mean Recall@K (mR@K) to evaluate unbiased SGG.
In this paper, we mainly discuss two overlooked issues by current metrics and propose two complementary metrics to evaluate unbiased SGG.
Different from mR@K, IMR treats each predicate category independently and WIMR assigns weights to different categories by taking the compositional diversity into consideration.

\section{Analyses of Current Unbiased SGG Benchmark}
In this section, we first detailed introduce the two overlooked issues in existing metrics and show their influence on evaluation results through data analysis. 
Then, we discuss the solutions to these issues and provide two complementary metrics for unbiased SGG evaluation.
  
\subsection{Category Independence}
The standard mR@K ranks all the predictions together by their estimated confidence scores to output top-K predictions, and then averages recall scores for each predicate category. It originally aims to treat each predicate category independently. 
However, we argue that ranking across categories in each image inadvertently involves cross-category interaction~\cite{dave2021evaluating}.
The current SGG task is formulated as a single-label classification problem and only one predicate is regarded as the ground truth. However, due to the strong label correlation in SGG dataset, there might be multiple reasonable predicates for one object pair. For example in Figure~\ref{fig:intro}(d), \texttt{looking-at}, \texttt{in-front-of} and \texttt{watching} are all reasonable predicates for the subject-object pair $\langle$\texttt{cat}, \texttt{laptop}$\rangle$. In such a case, the single-label annotation setting will bring unfairness for the predicates with high correlations when ranking across categories.
Consider a ``perfect'' SGG model which is expected to score all these possibly correct predicates high (\eg score 1.0 for all the above three predicates, 0.0 for the rest), the probability scores of these categories will be suppressed relatively after the normalization for single-label classification (\eg softmax).
Consequently, the recall score of the \texttt{watching} triplets will be underestimated as these predictions are less likely to appear in top-$K$ with low confidence scores.
Thus, it is unfair to compare and rank all predicates together.

\begin{figure}[t]
  \centering
    \vspace{-0.5em}
  \subfigure[Biased Outputs.]{
    \begin{minipage}{0.30\linewidth}
      \includegraphics[width=1.0\linewidth]{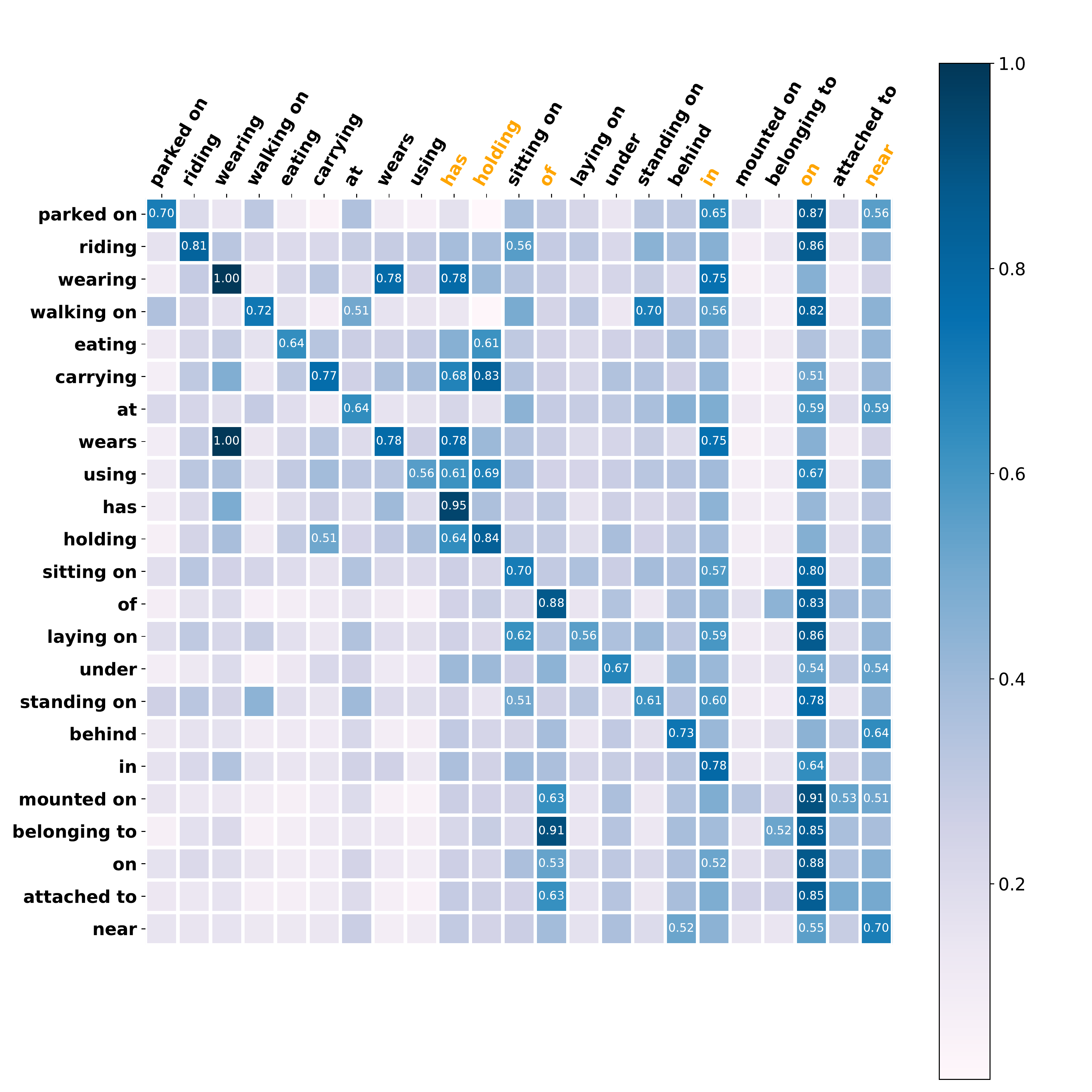}
    \end{minipage}
    \label{fig:biased}
    }
  \hfill
  \subfigure[Debiased Outputs.]{
    \begin{minipage}{0.30\linewidth}
      \includegraphics[width=1.0\linewidth]{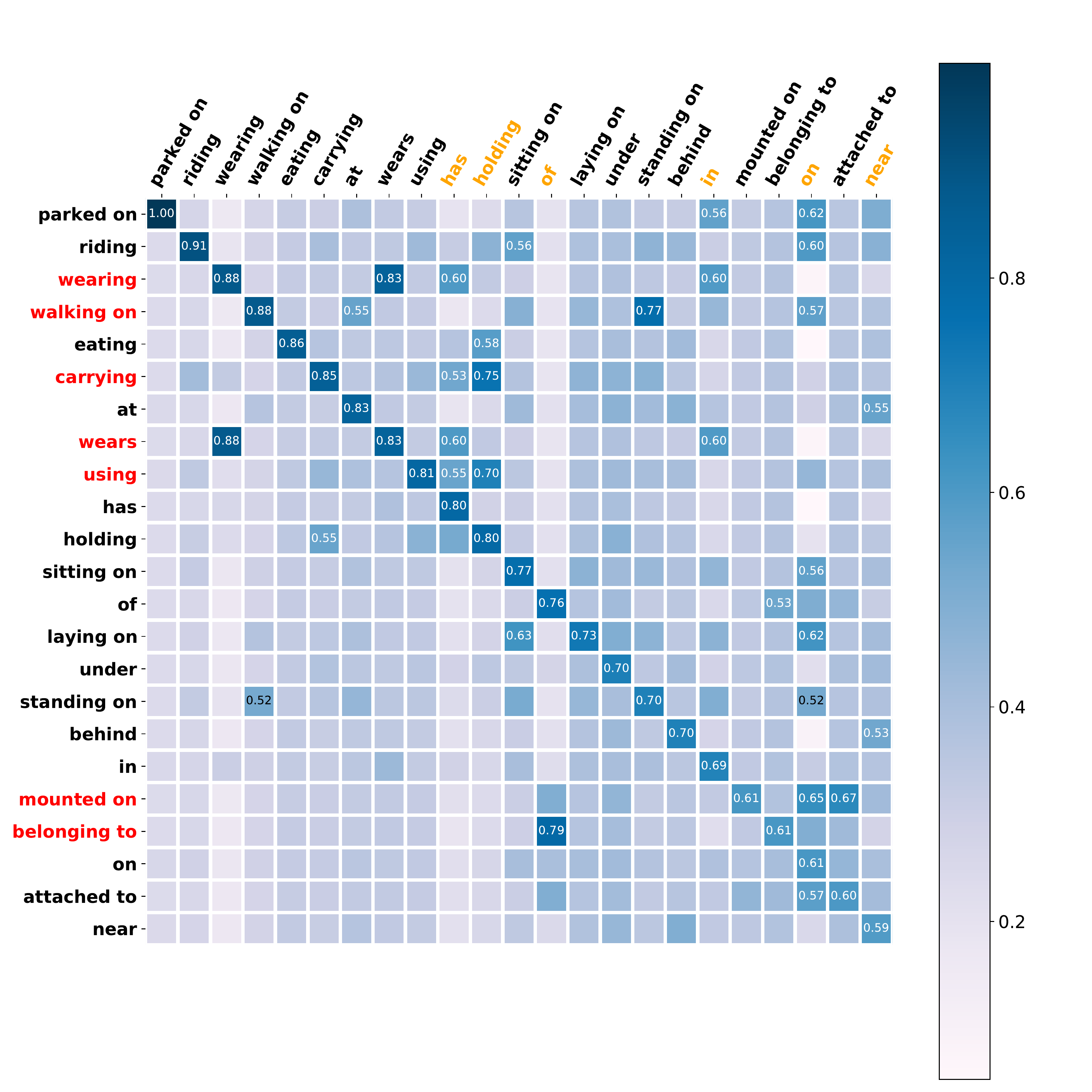}
      
    \end{minipage}
    \label{fig:unbiased}
    }
  \hfill
  \subfigure[Debiased Outputs after Softmax.]{
    \begin{minipage}{0.34\linewidth}
      \includegraphics[width=1.0\linewidth]{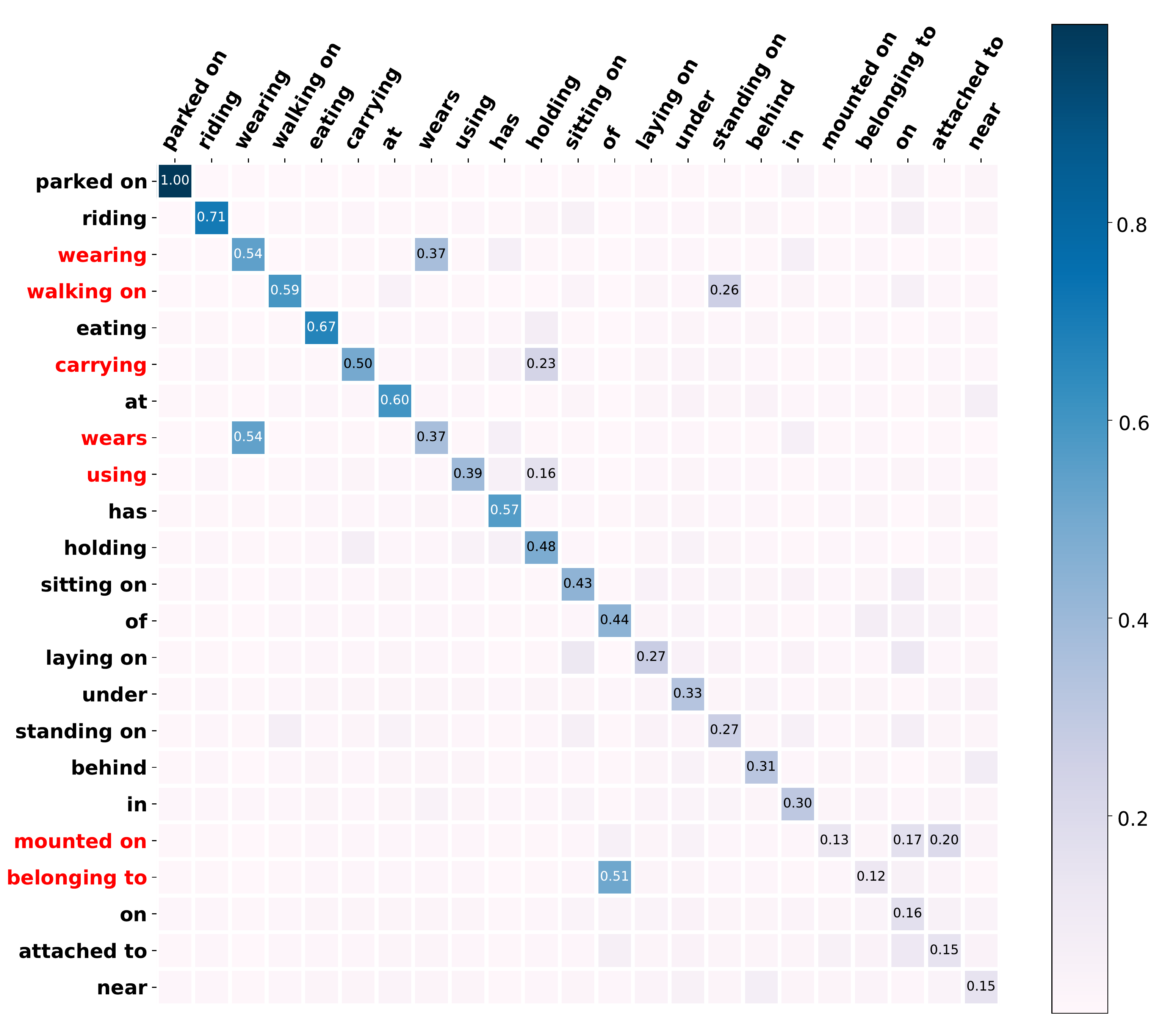}
    \end{minipage}
    \label{fig:softmax}
    }
  \caption{Visualization of the output for each predicate (ground truth). Each row of data in the heatmap represents the averaged output distribution of the samples annotated as the predicate left in the figure. For visualization, we normalize the scores over the whole map and only display a part of the predicates. }
  \label{fig:correlation}
\end{figure}

\noindent\textbf{Dataset Analysis.} To observe the influence of predicate correlations on evaluation by a real dataset, we visualize three output distributions for each predicate category based on a state-of-the-art SGG model Motifs~\cite{motifs} in Figure~\ref{fig:correlation}, including: (a) the logits distribution of biased Motifs model before softmax; (b) the logits distribution of unbiased Motifs model with reweight~\cite{kaihuaCausal} strategy\footnote{The inversed sample fractions were assigned to each predicate category as weights in cross-entropy loss.}; (c) the probability scores of the unbiased Motifs model after softmax.
All visualization outputs are averaged over the test set of VG.
% We average the output scores over all images in the dataset. \textcolor{red}{For visualization, we normalize the scores and only display a part of predicates}.
From the figure, we can easily observe two phenomena: 
(1) Comparing Figure~\ref{fig:biased} and Figure~\ref{fig:unbiased}, we can see that reweight strategy can alleviate the bias on the head predicates (\eg, \texttt{on} marked \textcolor{orange}{orange} in the figure) shown in Figure~\ref{fig:biased} to a certain extent. However, we can find there are some ubiquitous correlations among some predicate categories (marked \textcolor{red}{red} in the figure). For example, the samples annotated as \texttt{wearing} normally output high scores on its alternative predicates (\ie, \texttt{wears}, \texttt{has} and \texttt{in}) as well. Similarly, it also occurs between \texttt{walking on} and \texttt{standing on}, \texttt{belonging to} and \texttt{of}.
(2) Comparing Figure~\ref{fig:unbiased} and Figure~\ref{fig:softmax}, we can see that the normalized probabilities (used as confidence scores to rank) of the predicates (correctly recognized) with high correlations (\eg \texttt{wearing} and \texttt{belonging to}, marked \textcolor{red}{red} in the figure) are suppressed relatively compared to other predicates. In the subsequent ranking process, they are more likely to be sorted behind unfairly.
To this end, we suggest removing the setting of ranking across categories and sorting each category independently.

\subsection{Compositional Diversity}
Unbiased SGG aims to remove the impact of data imbalance in the model training process and encourage the models to recognize the tail predicates which have limited samples. Therefore, it assigns equal weights to each predicate category to reduce the proportion of head samples in performance. 
However, we argue that this category-wise averaging strategy can not reflect the rich structure of predicates on composed relationship triplets.
Hence, in this paper, we investigate the predicate categories from a new perspective: compositional diversity, \ie, the number of possible subject-object category pairs for each predicate.

\begin{figure}[t]
  \centering
\begin{minipage}{0.5\linewidth}
  \centering
  \setlength{\abovecaptionskip}{0.28cm}
  \includegraphics[width=\linewidth]{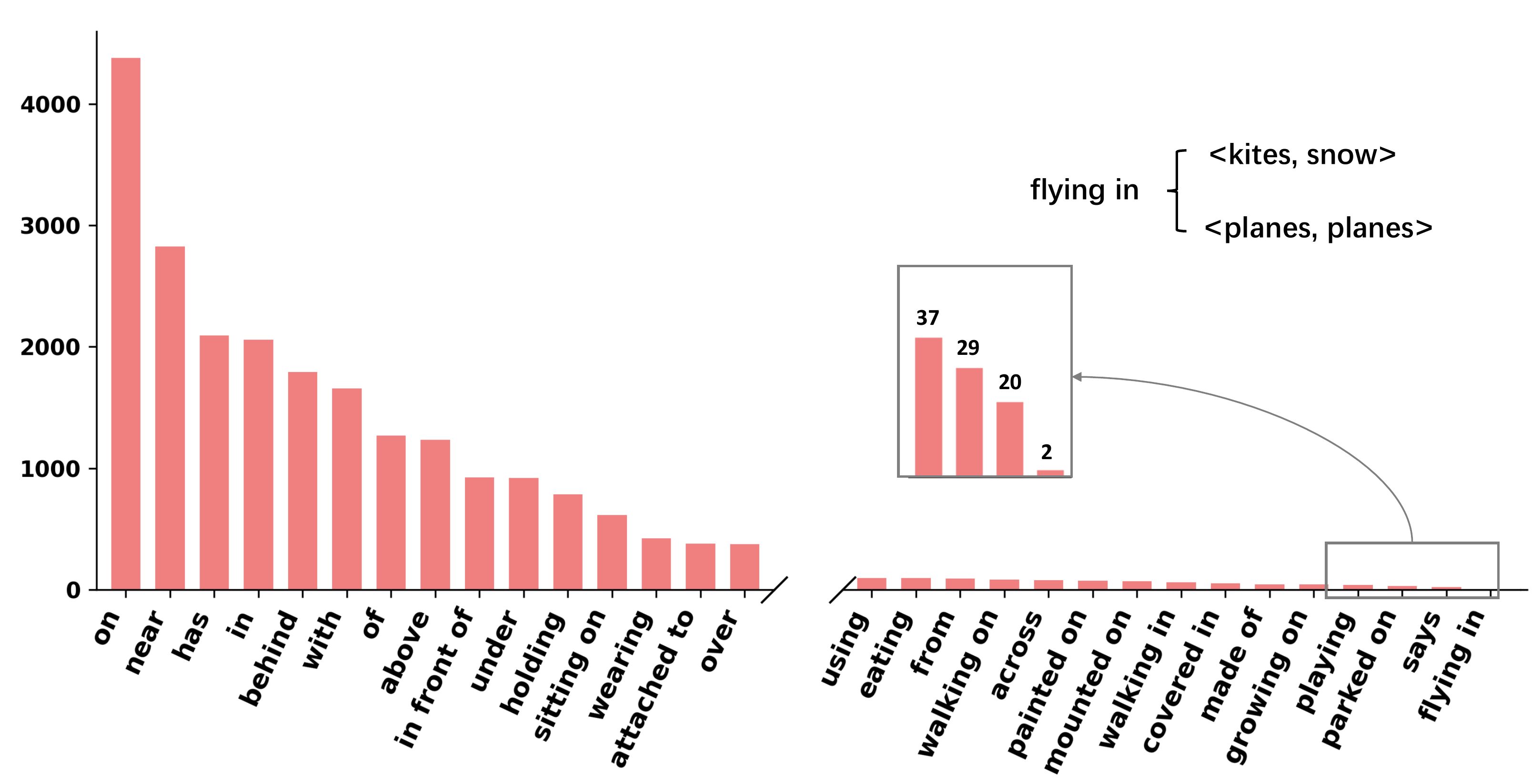}
  \vspace{-2em}
  \caption{The number of composed object pairs for each category in VG (30 predicates are shown).}
  \label{fig:semantic_diversity}
\end{minipage}
%\qquad
\hfill
\begin{minipage}{0.47\linewidth}
  \centering
      \scalebox{0.7}
          {
          \begin{tabular}{ |l|c|c | l| } 
              \hline
                & \#$\rm type_{pair}$ & mR@100 &\textbf{Impro.\%}  \\
              \hline
              Motifs (N=0) & - & 17.2 & - \\ 
            \hline
              $N=1$~\color{gray}{\small{(flying in)}}& 2 & 18.4 & \textbf{1.2} \\
            %   \hline
              $N=2$~\color{gray}{\small{(says)}} & 20 & 19.2 &  \textbf{2.0}~\color{gray}{+0.8}\\
            %   \hline
              $N=3$~\color{gray}{\small{(parked on)}}  & 29 & 21.0 & \textbf{3.8}~\color{gray}{+1.8} \\
            %   \hline
              $N=4$~\color{gray}{\small{(playing)}} & 37  & 20.9  & \textbf{3.7}~\color{gray}{-~0.1} \\
            %   \hline
              $N=5$~\color{gray}{\small{(growing on)}} & 41 & 22.3  & \textbf{5.1}~\color{gray}{+1.4} \\
            %   \hline
              $N=6$~\color{gray}{\small{(made of)}} & 42  & 22.9 & \textbf{5.7}~\color{gray}{+0.6}  \\
            %   \hline
              % $N=7$~\color{gray}{\small{(covered in)}} & 51  & 23.7 & \textbf{6.5}~\color{gray}{+0.8} \\
              \hline
              \end{tabular}
            }
    \vspace{-0.5em}
    \setlength{\abovecaptionskip}{0.28cm}
  %   \hspace{-0.5cm}
      \captionof{table}{Performance (\%) of Motifs~\cite{motifs} on PredCls by blindly replacing the results of the corresponding subject-object pairs for $N$ predicates with smallest compositional space. 
      % \#$\rm type_{pair}$ denotes the number of types of object pairs for the corresponding predicate. 
      }
      \label{tab:N}
    \end{minipage}
\end{figure}

\noindent\textbf{Studies of Compositional Diversity.}\label{sec:compositional_diversity}
To observe the compositional diversity of different predicates, 
we count the numbers of all possible composed subject-object category pairs for each predicate category in VG dataset. As shown in Figure~\ref{fig:semantic_diversity}, we can see that the compositional diversity of different predicates varies greatly, \ie,
the head categories in the figure typically have more generic semantics and can be applied to plenty of subject-object pairs, while the tail ones in Figure~\ref{fig:semantic_diversity} are more specific and can be only applied to a few subject-object category pairs. For example, predicate \texttt{parked on} mainly occurs between various vehicles and places, like $\langle$\texttt{car}, \texttt{parked-on}, \texttt{road}$\rangle$.

Meanwhile, we can observe the tail predicates (rare ones) in the SGG datasets usually possess limited compositional diversity. Thus, mR@K unconsciously assigns high weights to the samples of those predicates with limited compositional diversity. However, our experiments found that the performance of these tail predicates can be easily boosted by subject-object priors.
Here, we devised a simple experiment and recorded mR@K scores:
%under the following settings:
1) We firstly evaluate a well-trained state-of-the-art SGG model Motifs~\cite{motifs} on the PredCls\footnote{There are three typical SGG tasks: 1) \emph{Predicate Classification} (\textbf{PredCls}): Given the ground-truth of object bounding boxes (bboxes) and class labels, the model is required to predict the predicate class of pairwise objects. 2) \emph{Scene Graph Classification} (\textbf{SGCls}): Given the ground-truth of object bboxes, the model is required to predict both object classes and predicate classes of pairwise objects. 3) \emph{Scene Graph Detection} (\textbf{SGDet}): Given an image, the model is required to detect object bboxes, and predict object classes and predicate classes of pairwise objects.}.
2) For the last $N$ tail predicate categories in Figure~\ref{fig:semantic_diversity}, we find out all their possible composable subject-object pairs in the training dataset, then for each composable subject-object pair of predicate category $c$, we replace their predictions as the predicate $c$\footnote{For the same subject-object category pair with multiple tail predicate choices, we select the one with more limited compositional diversity.} at test. For example, when $N=1$, we directly predict the predicates of all subject-object pairs with categories $\langle$\texttt{kites}, \texttt{snow}$\rangle$ and  $\langle$\texttt{planes}, \texttt{planes}$\rangle$ as \texttt{flying-in}. We set $N$ from $1$ to $6$ and record mR@K performance in Table~\ref{tab:N}.
From the table, we can observe that the results on mR@K are continuously improved with the increase of $N$.
Specifically, mR@100 is improved by $1.2\%$ points only by blindly replacing the predictions for \texttt{flying-in}. Then, by further replacing the results of all corresponding subject-object pairs for predicate \texttt{says}, mR@100 is improved by $2.0\%$ points accumulatively.
\textbf{\emph{We can observe that the predicates with limited compositional diversity have a stronger correlation with subject-object priors, which can be simply improved even without visual information}}.

\subsection{Suggestion on Evaluation Metrics}
To handle the above-mentioned issues, we discuss and provide two complementary metrics to improve the current unbiased SGG evaluation benchmarking. In addition, we devise a simple baseline and improve unbiased SGG performance greatly.

\noindent\textbf{Independent Mean Recall (IMR).} Since there are actually many multiple reasonable predicates for some subject-object pairs, it is unfair to rank all triplet predictions across categories under the single-label classification problem formulation. One intuitive and straightforward solution to solve this problem is to build a well-annotated test dataset and redefine the task as a multi-label classification problem. However, it costs expensive to annotate exhaustively all the reasonable predicates for the subject-object pairs. Hence, we provide one solution to treat each predicate category independently and avoid their mutual influence.

For each image, we independently rank and output top-$K$ ($K=10/20/50$) predictions for each predicate category to calculate their own recall scores on this image. In this way, we remove the influence of different categories and the predicates with high correlations will not be suppressed by the other predicates. 
Then we obtain the recall on each predicate category by averaging the corresponding scores over all images. The final score is the averaged value over categories.
We call this metric Independent Mean Recall (IMR).

\begin{figure}
  \begin{minipage}[t]{0.48\linewidth}
  \begin{center}
      \includegraphics[width=\linewidth]{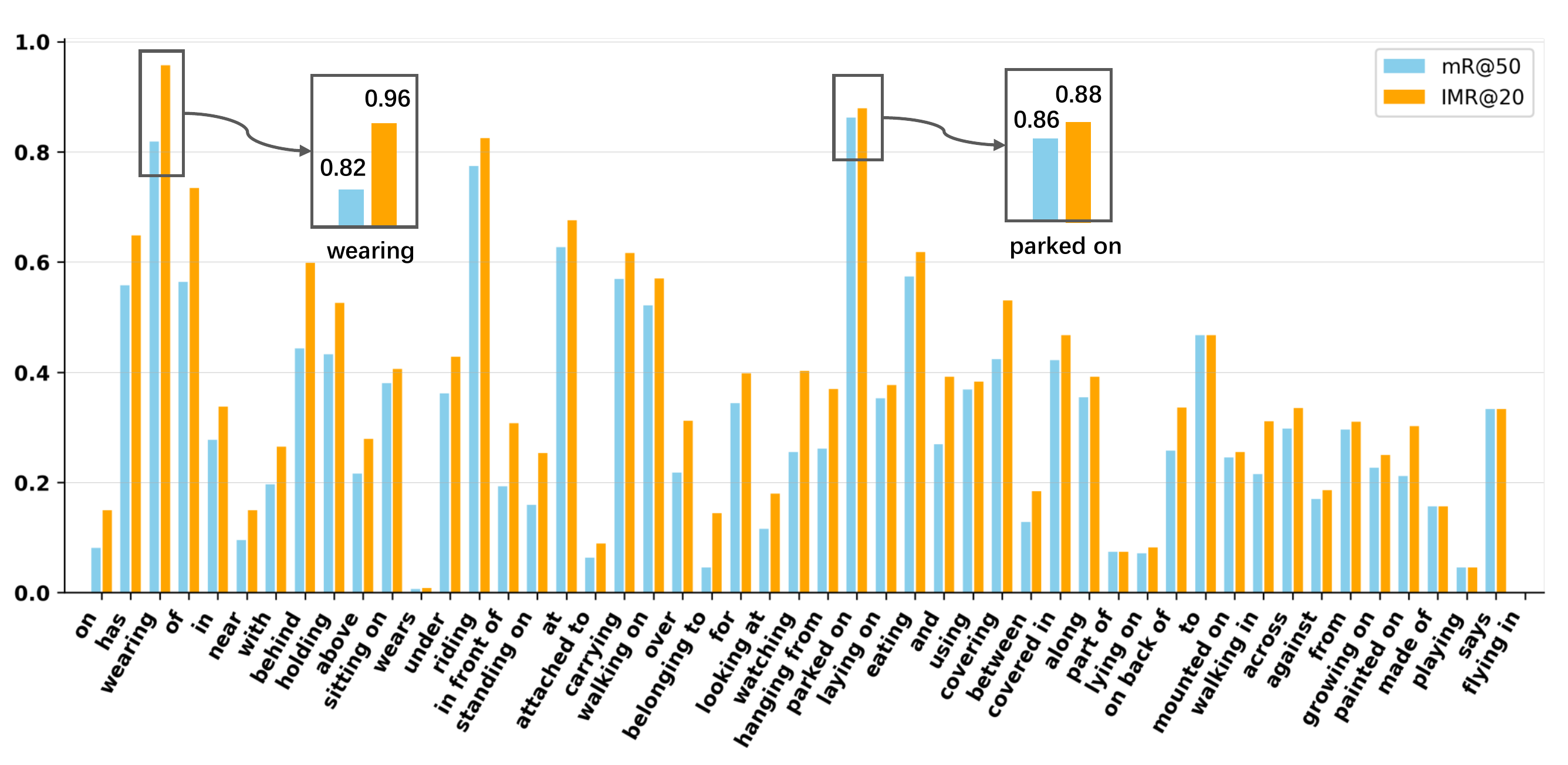}      
  \end{center}
  \vspace{-0.5em}
  \caption{Performance of each predicate category on mR@50 and IMR@20 for Motifs+Reweight (Predcls). Categories are displayed in descending order according to the number of samples in VG.}
  \label{fig:mR-ImR}
  \end{minipage}%
  \hfill
  \begin{minipage}[t]{0.48\linewidth}
  \begin{center}
    \includegraphics[width=\linewidth]{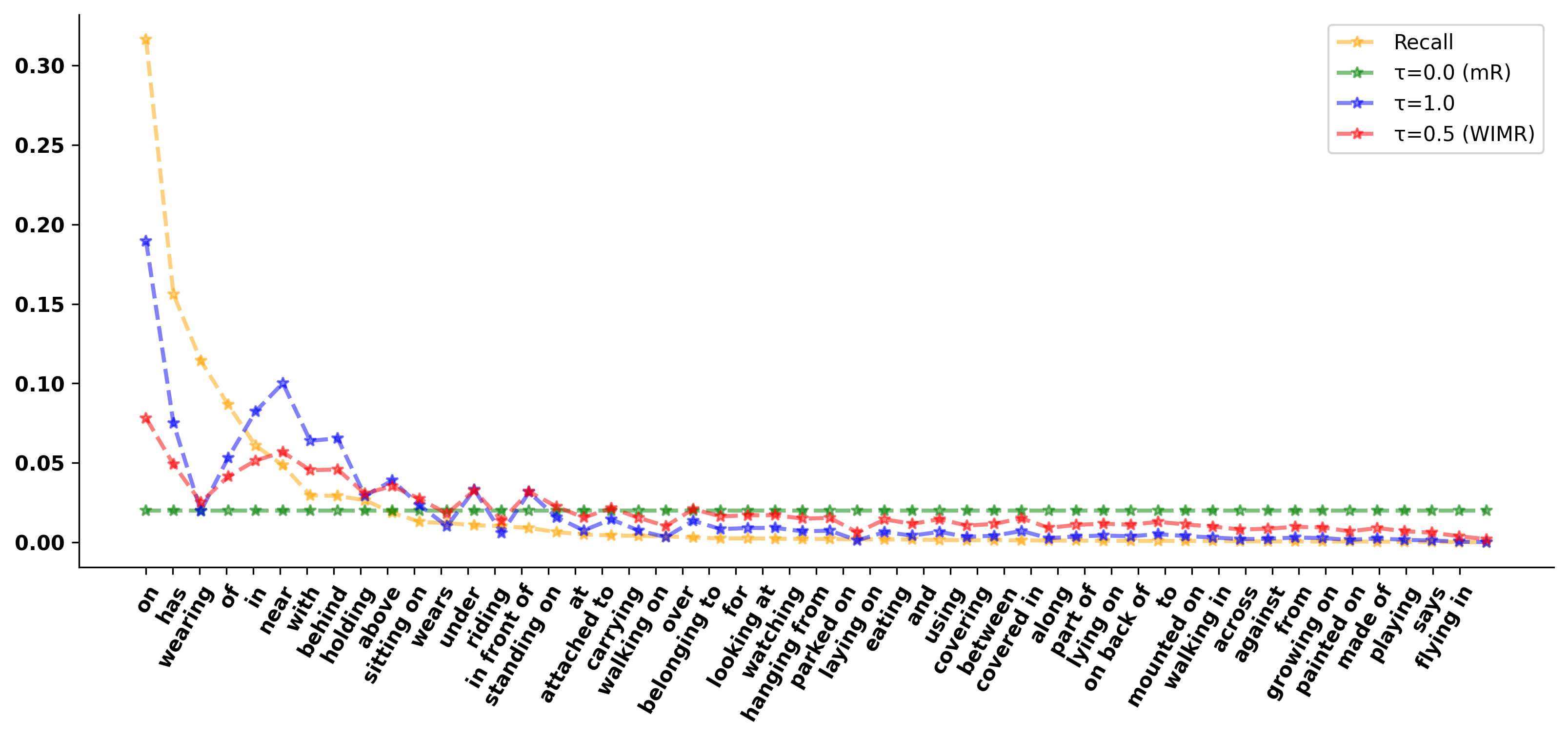}      
  \end{center}
  \vspace{-0.5em}
  \caption{Weights assigned to each predicate category on different metric settings. Categories are displayed in descending order according to the number of samples in VG.}
  \label{fig:weights}
  \end{minipage}
  \end{figure}

\noindent\textbf{Weighted Independent Mean Recall (wIMR).}
The standard mR@K treats each predicate category equally and assigns equal weights to each predicate category.
However, the target of the SGG task is to recognize more types of composed relation triplets rather than purely more kinds of predicates. Because the predicate categorical labels do not reflect the rich structures in the object relationships.
The composable diversity varies widely between different predicates.
Recent studies report the mean value of R@K and mR@K to expect models pay more attention to head predicates when realizing debiasing targets. However, R@K is not reliable because the variety of the number of head and tail predicates is too great. Besides, some predicates (\eg, \texttt{wearing}) with large distribution does not have rich semantics and have limited compositional diversity as shown in Figure~\ref{fig:weights}. 

Therefore, we suggest reassigning weights to each predicate category $c$ according to the complexity of their compositional space. 
We count the number of composed subject-object pairs $n_c$ for each predicate category, and reassign weights to each predicate category $c$: % The weight $w(n_c)$ is computed by:

\begin{equation}
  w(n_c)=\frac{n_c^{\tau }}{\sum_{k \in \mathcal{C} }n_k^{\tau }}, \qquad wIMR = \sum_{c \in \mathcal{C}} w(n_c) \times IMR(c),
\end{equation}
where $\mathcal{C}$ is the set of predicate categories, ${\tau} \in [0,1]$ controls the softness of weight distribution. When $\tau = 0$, wIMR equals IMR, which assigns equal weights to each predicate.

\section{Intrinsic Correlation between Predicates and Objects}
Different from other conventional scene understanding tasks, SGG predicts the predicates with corresponding subject-object pair information, which will bring in plenty of inductive information.
Zellers~\etal~\cite{motifs} finds the prediction of predicates is highly correlated with subject-object priors.
They count the co-occurrence of subjects, objects and predicates in relation triplets and obtain a statistics matrix from training dataset $\Mat{A} \in \mathbb{R}^{N_s\times N_o \times N_p}$ (where $N_s = N_o$ is the number of object categories, and $N_p$ is the number of predicate categories).
They aggregate the dimensions of subjects and objects to find the frequent predicate under the given pair.
For example, \texttt{on} is the most frequent predicate under $\langle$\texttt{car}, \texttt{street}$\rangle$.
However, this statistical prior actually collects the frequency bias and can not reflect the intrinsic correlation between predicates and objects. 
In this paper, we find that it is more reasonable to observe the distribution of objects over each predicate to find the correlation between predicates and objects. 
For example, if we count the most frequent objects given \texttt{parked on}, we can find its most relevant object category is \texttt{car}, which is in line with common sense. 
We call this statistical prior \emph{Predicate Knowledge on Objects (\textbf{PKO})}.
Reasonably exploiting this part of statistical knowledge PKO can improve unbiased SGG models' performance.

In this paper, we devise a simple baseline to directly aggregate this statistical prior into the inference results of SGG models following the frequency prior in Motifs~\cite{motifs}.
From $\Mat{A}$, we can easily obtain the predicate-subject co-occurrence matrix $\Mat{A}^s \in \mathbb{R}^{N_p\times N_s}$ and predicate-object co-occurrence matrix $\Mat{A}^o \in \mathbb{R}^{N_p\times N_o}$, separately.
Then, we normalize the matrices on the dimension of subject (object) and obtain the distribution of subject (object) under all predicates: $\tilde{\Mat{A}}^s \in \mathbb{R}^{N_p \times N_s}$ and $\tilde{\Mat{A}}^o \in \mathbb{R}^{N_p\times N_o}$.
Finally, given a subject-object pair $(i,j)$, an SGG model will output the predicate logits $\hat{\Mat{z}}_{i,j} \in \mathbb{R}^{N_p}$, we aggregate the statistical prior by:
\begin{equation}
  \Mat{z}_{i,j} = \hat{\Mat{z}}_{i,j} +\Mat{b}_{i,j}, \qquad 
  b_{i,j,k} = - \text{log} \frac{\tilde{\Mat{A}}^s_{k,i}}{ \textstyle \sum_{c\in \mathcal{C} } \tilde{\Mat{A}}^s_{c,i} } - \text{log} \frac{\tilde{\Mat{A}}^o_{k,j}}{ \textstyle \sum_{c\in \mathcal{C} }\tilde{\Mat{A}}^o_{c,j}},
\end{equation}
where $\mathcal{C} $ is the set of predicate categories and $\Mat{z}_{i,j}$ is the final output of the SGG model. We can find this simple baseline improves unbiased performance greatly (in Sec.~\ref{sec:exp}). The main improvements are on the predicates with limited compositional diversity which verifies the statements in Sec.~\ref{sec:compositional_diversity}, \ie, these predicates are more highly correlated with subject-object priors and can be simply improved even without visual information.

\section{Experiments}
In this section, we develop experiments to discuss the proposed metrics and compare them with current metrics. We also evaluate different unbiased SGG methods on these metrics.

% \subsection{Experimental settings}
\paragraph{Dataset.}
We evaluated all results on the challenging Visual Genome~\cite{visualGenome}, which is a large-scale widely-used benchmark in SGG tasks. We adopt the popular split~\cite{motifs, PCPL}, which includes the most frequent 150 object categories and 50 predicate classes.
After preprocessing, each image has $11.5$ objects and $6.2$ relationships on average. The split uses $70\%$ of images for training and $30\%$ for the test.

\subsection{Comparison between Existing and New Metrics}

\noindent\textbf{IMR@K \emph{vs.} mR@K.}~\label{sec:imr_exp}
% \paragraph{IMR@K \emph{vs.} mR@K.}~\label{sec:imr_exp}
In Figure~\ref{fig:mR-ImR}, we display the recall scores of each predicate category on mR@50 and IMR@20 for Motifs-Reweight \wrt~PredCls, respectively.
We can see that the recall scores of the predicates in Figure~\ref{fig:correlation} underestimated by mR@K are more fairly evaluated on IMR@K.
For example, the recall scores of the high correlated predicates, like \texttt{wearing}, \texttt{of} and \texttt{watching}, ascend significantly on IMR@20 compared to mR@50. While the predicate \texttt{parked on} has no obvious improvement on IMR@20 as it is slightly influenced by the confidence sharing shown in Figure~\ref{fig:softmax}. Compared to mR@K, IMR@K provides a more fair score for these over-estimated predicate categories.

\noindent\textbf{wIMR@K \emph{vs.} mR@K/R@K.}
% \paragraph{wIMR@K \emph{vs.} mR@K/R@K.}
In Figure~\ref{fig:weights}, we display the weights of different predicates (on VG dataset) in R@K, mR@K/IMR ($\tau=0.0$) and wIMR ($\tau=0.5$).
We can observe that the classic R@K mainly focuses on head predicates although some head predicates have limited compositional diversity (\eg \texttt{wearing}).
% To prevent SGG models from biasing towards the head predicates, 
While mR@K assigns equal weights to all predicates despite their difference in compositional diversity, and the samples of some predicates with limited compositional diveristy like \texttt{flying-in} are assigned excessively high weights. In contrast, wIMR, by considering compositional diversity, manages to assign high weights to predicates with rich semantics (\eg,\texttt{of} and \texttt{under}) and low weights to predicates with simple semantics (\eg,\texttt{flying-in} and \texttt{wearing}).
The value of $\tau$ controls the softness of the weight distribution and can be adjusted according to task target, we set $\tau=0.5$ in the default implementation of wIMR.

\begin{table*}
  \centering
  \scalebox{0.7}
  {
  \begin{tabular}{|r | r | c c c c | }
  \hline
  \multirow{2}{*}{} & \multirow{2}{*}{Method} & \multicolumn{4}{c|}{PredCls} \\
  & & R@20~/~50~/~100 & mR@20~/~50~/~100 & IMR@10~/~20~/~50 & wIMR@10~/~20~/~50 \\

  \hline
  & \cellcolor{mygray-bg}{\textbf{PKO}} & \cellcolor{mygray-bg}{10.10~/~15.51~/~19.44}  & \cellcolor{mygray-bg}{17.86~/~27.38~/~33.93}  & \cellcolor{mygray-bg}{\textcolor{red}{\textbf{38.03}}~/~\textcolor{red}{\textbf{39.02}}~/~\textcolor{red}{\textbf{40.90}}} & \cellcolor{mygray-bg}{29.75~/~30.79~/~31.38}\\
  \hline
  \multirow{5}*{Motifs} & Baseline & \textcolor{red}{\textbf{59.19}}~/~\textcolor{red}{\textbf{65.59}}~/~\textcolor{red}{\textbf{67.30}} & 12.62~/~15.96~/~17.23 & 14.53~/~16.11~/~17.45 & 24.80~/~27.99~/~30.97 \\ 
  & TDE & 33.35~/~45.89~/~51.24 & 17.85~/~24.77~/~28.72 & 27.33~/~29.52~/~31.06 & 32.64~/~35.59~/~37.72 \\
  % & Naive (N=11) & & & & \\
  & DLFE & 44.28~/~50.33~/~51.99 & 21.86~/~26.81~/~28.53 & 26.37~/~27.92~/~29.03 &  31.49~/~34.02~/~35.73\\
  & NICE & 48.15~/~55.14~/~57.15 & 23.67~/~29.83~/~32.24 & 28.35~/~31.31~/~33.12  & 31.61~/~35.26~/~37.84 \\
  & Reweight & 26.57~/~36.08~/~40.39 & \textcolor{blue}{23.98}~/~\textcolor{blue}{30.79}~/~\textcolor{blue}{34.48} & \textcolor{red}{\textbf{35.39}}~/~\textcolor{red}{\textbf{36.58}}~/~\textcolor{red}{\textbf{36.98}} &  \textcolor{red}{\textbf{36.13}}~/~\textcolor{red}{\textbf{37.52}}~/~38.05 \\
  & RTPB(CB) & 36.58~/~42.64~/~44.36 & \textcolor{red}{\textbf{27.61}}~/~\textcolor{red}{\textbf{32.78}}~/~\textcolor{red}{\textbf{34.57}} & \textcolor{blue}{30.55}~/~\textcolor{blue}{33.30}~/~\textcolor{blue}{35.02} & \textcolor{blue}{33.63}~/~\textcolor{blue}{37.16}~/~\textcolor{red}{\textbf{39.32}} \\
  &\cellcolor{mygray-bg}{\textbf{PKO}} & \cellcolor{mygray-bg}{\textcolor{blue}{49.08}~/~\textcolor{blue}{55.95}~/~\textcolor{blue}{58.18}}  & \cellcolor{mygray-bg}{24.98~/~31.44~/~33.98}  & \cellcolor{mygray-bg}{29.73~/~32.46~/~34.21} & \cellcolor{mygray-bg}{31.77~/~35.51~/~\textcolor{blue}{38.51}} \\
  
  \hline
  \multirow{5}*{VCTree} & Baseline & \textcolor{red}{\textbf{59.72}}~/~\textcolor{red}{\textbf{65.86}}~/~\textcolor{red}{\textbf{67.50}} & 13.26~/~16.82~/~18.12 & 15.36~/~16.99~/~18.30  & 25.31~/~28.58~/~31.58 \\ 
  & TDE & 34.48~/~44.89~/~49.20 & 19.07~/~25.61~/~29.13 & 27.24~/~29.46~/~31.03 & 32.59~/~35.70~/~38.05 \\
  % & Naive (N=11) & & & & \\
  & DLFE & 45.35~/~51.21~/~52.75  & 22.53~/~27.36~/~28.86  & 26.46~/~28.28~/~29.30 & 31.49~/~34.27~/~36.01 \\
  & NICE & 48.38~/~55.03~/~56.92 & 24.42~/~30.74~/~33.01 & 29.03~/~32.01~/~33.93 & 31.72~/~35.41~/~38.12 \\
  & Reweight & 28.66~/~35.62~/~37.90 & \textcolor{red}{\textbf{28.64}}~/~\textcolor{red}{\textbf{34.93}}~/~\textcolor{red}{\textbf{37.28}} & \textcolor{red}{\textbf{34.70}}~/~\textcolor{red}{\textbf{36.95}}~/~\textcolor{red}{\textbf{38.30}} & \textcolor{red}{\textbf{34.41}}~/~\textcolor{blue}{36.94}~/~38.42 \\
  & RTPB(CB) & 36.65~/~42.39~/~43.95 & \textcolor{blue}{28.64}~/~\textcolor{blue}{33.41}~/~\textcolor{blue}{35.11} & 30.57~/~\textcolor{blue}{33.47}~/~33.50 & \textcolor{blue}{33.22}~/~\textcolor{red}{\textbf{36.99}}~/~\textcolor{red}{\textbf{39.50}} \\
  & \cellcolor{mygray-bg}{\textbf{PKO}} & \cellcolor{mygray-bg}{\textcolor{blue}{49.39}~/~\textcolor{blue}{56.06}~/~\textcolor{blue}{58.18}}  & \cellcolor{mygray-bg}{26.06~/~32.20~/~34.61 } & \cellcolor{mygray-bg}{\textcolor{blue}{30.61}~/~33.41~/~\textcolor{blue}{35.29}} & \cellcolor{mygray-bg}{31.98~/~35.73~/~\textcolor{blue}{38.88}} \\
  \hline 

  \end{tabular}
  }
  \setlength{\abovecaptionskip}{0.1cm}
  \caption{Performance comparison of different unbiased SGG methods (\wrt Predcls) on metrics: R@K, mR@K, IMR@K, wIMR@K.
  VCTree~\cite{vcTree} and Motifs~\cite{motifs} are two popular compared baseline models in unbiased SGG. All experimental results are re-implemented using official released codes. 
% The experimental results of SGCls and SGDet are reported in supplementary material. 
  The \textcolor{red}{\textbf{best}} and \textcolor{blue}{second best} results are marked according to formats. }
  \label{tab:mR-IMR}
\end{table*}

\subsection{Evaluation on Current Methods}~\label{sec:exp}
\noindent\textbf{Benchmarking SOTA Unbiased SGG Methods.}
For the convenience to further compare with current unbiased SGG methods, we report the performance of five SOTA unbiased SGG methods on current and new proposed metrics, including \emph{TDE}~\cite{kaihuaCausal}, \emph{DLFE}~\cite{chiou2021recovering}, \emph{NICE}~\cite{li2022devil}, \emph{RTPB(CB)}~\cite{chen2022resistance} and \emph{Reweight}~\cite{kaihuaCausal}.
The details are reported in Table~\ref{tab:mR-IMR} and Table~\ref{tab:sgcls-sgdet}.

\noindent\textbf{Effectiveness of Predicate Knowledge on Objects (PKO).}
We also report the performance of models with our proposed statistical prior in Table~\ref{tab:mR-IMR} and Table~\ref{tab:sgcls-sgdet}.
Wherein, the simple baseline, \emph{PKO}, means that we only utilize the statistical prior to make predictions. Amazingly, we can find it achieves excellent performance on IMR@K and mR@K. However, as it mainly improves the performance of the predicates with limited compositional diversity, it did not perform well on wIMR@K and R@K.
We aggregate PKO into other SGG models, and it greatly improves their unbiased performance (\eg \emph{Motifs-PKO} and \emph{VCTree-PKO}).

\begin{table*}[t]
    \centering
    \scalebox{0.7}
    {
    \begin{tabular}{|r | r | c c c c | }
    \hline
    \multirow{2}{*}{} & \multirow{2}{*}{Method} & \multicolumn{4}{c|}{SGCls} \\
    & & R@20~/~50~/~100 & mR@20~/~50~/~100 & IMR@10~/~20~/~50 & wIMR@10~/~20~/~50 \\
  
    \hline
    & \cellcolor{mygray-bg}{\textbf{PKO}} & \cellcolor{mygray-bg}{7.26~/~10.48~/~12.46 }  & \cellcolor{mygray-bg}{11.23~/~16.77~/~20.13 }  & \cellcolor{mygray-bg}{20.51~/~22.47~/~23.11} & \cellcolor{mygray-bg}{16.04~/~17.13~/~17.68 }\\
    \hline
    \multirow{5}*{Motifs} & Baseline  & \textbf{\textcolor{red}{36.39}}~/~\textbf{\textcolor{red}{39.59}}~/~\textbf{\textcolor{red}{40.35}} & 7.44~/~9.09~/~9.63 &  7.69~/~8.65~/~9.57 & 13.65~/~15.70~/~17.76 \\ 
    & TDE & 20.46~/~26.31~/~28.78 & 9.78~/~13.21~/~15.07  & 13.98~/~15.22~/~16.29   &  16.95~/~18.69~/~20.16 \\
    % & Naive (N=11) & & & & \\
    & DLFE & 26.63~/~29.79~/~30.61  & 13.23~/~15.66~/~16.48  & 14.75~/~15.82~/~16.57  & 17.96~/~19.65~/~20.85  \\
    & NICE & 29.48~/~33.06~/~34.05 & 13.63~/~16.67~/~17.88  & 14.82~/~16.76~/~18.02  & 17.00~/~19.35~/~21.07   \\
    & Reweight & 18.67~/~23.49~/~25.47   & 13.49~/~16.75~/~18.34  & \textbf{\textcolor{red}{17.82}}~/~\textbf{\textcolor{red}{18.69}}~/~\textbf{\textcolor{red}{19.23}} & \textbf{\textcolor{red}{19.13}}~/~\textcolor{blue}{20.38}~/~21.14 \\
    & RTPB(CB) & 22.66~/~25.85~/~26.65  & \textbf{\textcolor{red}{15.77}}~/~\textbf{\textcolor{red}{18.16}}~/~\textcolor{blue}{18.97}  & \textcolor{blue}{16.08}~/~\textcolor{blue}{17.68}~/~\textcolor{blue}{18.81}  & 18.51~/~\textbf{\textcolor{red}{20.61}}~/~\textbf{\textcolor{red}{21.99}}   \\
    &\cellcolor{mygray-bg}{\textbf{PKO}} & \cellcolor{mygray-bg}{\textcolor{blue}{30.41}~/~\textcolor{blue}{33.99}~/~\textcolor{blue}{35.05}}  & \cellcolor{mygray-bg}{\textcolor{blue}{14.06}~/~\textcolor{blue}{17.59}~/~\textbf{\textcolor{red}{19.12}}}  & \cellcolor{mygray-bg}{15.87~/~17.49~/~18.66} & \cellcolor{mygray-bg}{ 17.43~/~19.72~/~\textcolor{blue}{21.72}} \\
    
    \hline
    \multirow{5}*{VCTree} & Baseline & \textbf{\textcolor{red}{42.09}}~/~\textbf{\textcolor{red}{45.80}}~/~\textbf{\textcolor{red}{46.73}} & 9.09~/~11.28~/~12.04  & 9.72~/~10.89~/~11.94 & 16.52~/~18.90~/~21.26  \\ 
    & TDE  & 23.48~/~31.17~/~34.59  & 10.36~/~14.47~/~16.72  & 15.81~/~17.32~/~18.48  & 19.71~/~21.73~/~23.35  \\
    % & Naive (N=11) & & & & \\
    & DLFE  & 30.09~/~33.85~/~34.80   & 16.17~/~19.23~/~20.20   & 18.81~/~20.04~/~20.62  & 21.75~/~23.58~/~24.65  \\
    & NICE & 33.77~/~37.84~/~38.99 & 16.14~/~20.03~/~21.29 & 17.89~/~19.91~/~21.59 & 20.16~/~22.77~/~24.80 \\
    & Reweight  & 20.44~/~24.66~/~25.97 & \textbf{\textcolor{red}{18.72}}~/~\textbf{\textcolor{red}{22.88}}~/~\textbf{\textcolor{red}{24.19}}   & \textbf{\textcolor{red}{22.14}}~/~\textbf{\textcolor{red}{23.66}}~/~\textbf{\textcolor{red}{24.58}}  &  \textcolor{blue}{22.04}~/~\textcolor{blue}{23.93}~/~25.07 \\
    & RTPB(CB)  & 26.06~/~29.63~/~30.61  & \textcolor{blue}{18.67}~/~21.41~/~22.52  & 19.64~/~21.41~/~22.63  & \textbf{\textcolor{red}{22.04}}~/~\textbf{\textcolor{red}{24.41}}~/~\textcolor{blue}{25.93}  \\
    & \cellcolor{mygray-bg}{\textbf{PKO}} & \cellcolor{mygray-bg}{\textcolor{blue}{35.01}~/~\textcolor{blue}{39.10}~/~\textcolor{blue}{40.40} }  & \cellcolor{mygray-bg}{ 18.43~/~\textcolor{blue}{22.27}~/~\textcolor{blue}{23.74} } & \cellcolor{mygray-bg}{\textcolor{blue}{19.88}~/~\textcolor{blue}{21.72}~/~\textcolor{blue}{23.01}} & \cellcolor{mygray-bg}{21.14~/~23.77~/~\textbf{\textcolor{red}{26.09}}} \\
    \hline 

    \hline
    \multirow{2}{*}{} & \multirow{2}{*}{Method} & \multicolumn{4}{c|}{SGDet} \\
    & & R@20~/~50~/~100 & mR@20~/~50~/~100 & IMR@10~/~20~/~50 & wIMR@10~/~20~/~50 \\
  
    \hline
    & \cellcolor{mygray-bg}{\textbf{PKO}} & \cellcolor{mygray-bg}{ 5.53~/~7.77~/~9.46 }  & \cellcolor{mygray-bg}{7.48~/~10.97~/~13.85 }  & \cellcolor{mygray-bg}{ \textbf{\textcolor{red}{19.37}}~/~\textbf{\textcolor{red}{19.87}}~/~\textbf{\textcolor{red}{20.76}}} & \cellcolor{mygray-bg}{15.17~/~15.48~/~16.14}\\
    \hline
    \multirow{5}*{Motifs} & Baseline  & \textbf{\textcolor{red}{25.79}}~/~\textbf{\textcolor{red}{33.04}}~/~\textbf{\textcolor{red}{37.49}} & 5.37~/~7.37~/~8.61  & 7.99~/~8.33~/~8.85  &  14.98~/~15.26~/~15.80 \\ 
    & TDE  & 11.94~/~16.57~/~20.15  & 6.54~/~8.93~/~10.95  & 13.67~/~13.89~/~14.55  &  \textcolor{blue}{17.58}~/~17.72~/~18.18 \\
    % & Naive (N=11) & & & & \\
    & DLFE  & 18.29~/~24.52~/~28.45  & 8.50~/~11.29~/~13.24 & 12.15~/~13.18~/~14.58  & 16.12~/~16.88~/~18.09  \\
    & NICE & \textcolor{blue}{21.30}~/~\textcolor{blue}{27.83}~/~\textcolor{blue}{31.75} & 8.82~/~12.23~/~14.40  & 12.89~/~13.42~/~14.41  & 16.54~/~16.96~/~17.78  \\
    & Reweight  & 12.63~/~17.26~/~20.59  & 8.88~/~11.78~/~14.09  & \textbf{\textcolor{red}{15.36}}~/~\textbf{\textcolor{red}{16.12}}~/~\textbf{\textcolor{red}{17.75}} & \textbf{\textcolor{red}{17.71}}~/~\textbf{\textcolor{red}{18.20}}~/~\textbf{\textcolor{red}{19.31}}  \\
    & RTPB(CB)  & 14.72~/~20.13~/~23.75  & \textbf{\textcolor{red}{11.35}}~/~\textbf{\textcolor{red}{14.45}}~/~\textbf{\textcolor{red}{16.74}}  & 13.60~/~14.35~/~15.91   & 16.65~/~17.15~/~18.31  \\
    &\cellcolor{mygray-bg}{\textbf{PKO}} & \cellcolor{mygray-bg}{ 20.38~/~26.95~/~31.14}  & \cellcolor{mygray-bg}{\textcolor{blue}{9.64}~/~\textcolor{blue}{13.44}~/~\textcolor{blue}{16.06} }  & \cellcolor{mygray-bg}{\textcolor{blue}{14.66}~/~\textcolor{blue}{15.39}~/~\textcolor{blue}{16.82} } & \cellcolor{mygray-bg}{17.48~/~\textcolor{blue}{17.96}~/~\textcolor{blue}{18.97}} \\
    
    \hline
    \multirow{5}*{VCTree} & Baseline & \textbf{\textcolor{red}{24.96}}~/~\textbf{\textcolor{red}{32.15}}~/~\textbf{\textcolor{red}{36.36}} & 5.20~/~7.05~/~8.28  & 7.44~/~7.68~/~8.11 & 14.34~/~14.53~/~14.98  \\ 
    & TDE  & 12.11~/~17.07~/~20.72  & 6.73~/~9.32~/~11.28  & \textcolor{blue}{13.57}~/~13.73~/~14.28  & \textcolor{blue}{17.69}~/~\textcolor{blue}{17.81}~/~\textcolor{blue}{18.17}  \\
    % & Naive (N=11) & & & & \\
    & DLFE & 17.61~/~23.58~/~27.21   & 8.76~/~11.63~/~13.32   & 11.80~/~12.73~/~14.24 & 15.03~/~15.68~/~16.87   \\
    & NICE & \textcolor{blue}{20.70}~/~\textcolor{blue}{26.99}~/~\textcolor{blue}{30.78}  & 8.39~/~11.96~/~14.08 & 12.84~/~13.28~/~14.27  & 16.13~/~16.49~/~17.30  \\
    & Reweight & 15.35~/~20.30~/~23.72  &  \textcolor{blue}{9.90}~/~12.69~/~15.00 & 13.23~/~\textcolor{blue}{14.08}~/~\textcolor{blue}{15.80}  & 15.98~/~16.50~/~17.73 \\
    & RTPB(CB)  & 14.52~/~19.70~/~23.17  & \textbf{\textcolor{red}{10.79}}~/~\textbf{\textcolor{red}{13.68}~/~\textbf{\textcolor{red}{15.85}}}  & 12.89~/~13.51~/~15.12   & 15.95~/~16.38~/~17.59 \\
    & \cellcolor{mygray-bg}{\textbf{PKO}} & \cellcolor{mygray-bg}{20.00~/~26.48~/~30.68 }  & \cellcolor{mygray-bg}{ 9.59~/~\textcolor{blue}{13.24}~/~\textcolor{blue}{15.85} } & \cellcolor{mygray-bg}{\textbf{\textcolor{red}{15.35}}~/~\textbf{\textcolor{red}{15.85}}~/~\textbf{\textcolor{red}{17.25 }}} & \cellcolor{mygray-bg}{\textbf{\textcolor{red}{17.77}}~/~\textbf{\textcolor{red}{18.11}}~/~\textbf{\textcolor{red}{19.05}}} \\
    \hline 
  
    \end{tabular}
    }
    \setlength{\abovecaptionskip}{0.1cm}
    \caption{Performance on SGCls and SGDet.
    % VCTree~\cite{vcTree} and Motifs~\cite{motifs} are two popular compared baseline models in unbiased SGG.
    All experimental results are re-implemented using official released codes. 
    The \textcolor{red}{\textbf{best}} and \textcolor{blue}{second best} results are marked according to formats.}
    \label{tab:sgcls-sgdet}
  \end{table*}

\section{Conclusions} 
% The evaluation of unbiased SGG models is non-trivial due to some inevitable annotation characteristics of SGG datasets.
In this paper, we focused on two overlooked issues which make the current evaluation benchmark vulnerable and unfair:
1) mR@K unintentionally breaks the category independence when ranking across categories;
2) mR@K assigns equal weights for each predicate but neglects the compositional diversity of subject-object pairs.
We discussed the influence of these issues through statistical data analysis and provided two complementary metrics for unbiased SGG evaluation to address the two above-mentioned issues.
We additionally investigated a statistical prior which reflects the intrinsic correlation between predicates and objects. Based on this, we devised a simple but strong baseline for unbiased SGG research.
% And to address the two above-mentioned issues, we introduced two complementary metrics for unbiased SGG evaluation.
We hope that our work is able to contribute some insights to this area and help benchmark unbiased SGG methods in a more trustworthy way.
Furthermore, our discussion is general which may be extended to video SGG scenarios~\cite{gao2022classification,gao2021video}.

\subsection*{Acknowledgements}
\begin{sloppypar}
This work was supported by the National Key Research \& Development Project of China (2021ZD0110700), the National Natural Science Foundation of China (U19B2043, 61976185), Zhejiang Natural Science Foundation (LR19F020002), Zhejiang Innovation Foundation (2019R52002), and the Fundamental Research Funds for the Central Universities (226-2022-00087).
\end{sloppypar}

\bibliography{reference}
\end{document}